%% file: root.tex
\documentclass[letterpaper, 10 pt, conference]{ieeeconf}  

\IEEEoverridecommandlockouts                              

\usepackage{booktabs}
\usepackage{multicol}
\usepackage{graphicx}
\usepackage{graphics} 
\usepackage{epsfig} 
\usepackage{amsmath} 
\usepackage{amssymb}
\usepackage{booktabs}
\usepackage{multirow}
\usepackage{algpseudocode}
\usepackage{algorithm}
\usepackage[algo2e]{algorithm2e}
\usepackage[normalem]{ulem}
\usepackage{gensymb}
\usepackage{color}
\usepackage[font=small,labelfont=bf]{caption} 
\newcommand{\ours}{DeformNet~}
\usepackage{balance}
\usepackage{cite}

\title{\LARGE \bf
DeformNet: Latent Space Modeling and Dynamics Prediction for Deformable Object Manipulation
}

\author{Chenchang Li$^{\dag}$, 
        Zihao Ai$^{\dag}$,
        Tong Wu$^{}$,
        Xiaosa Li$^{}$,
        Wenbo Ding$^{*}$,
        and Huazhe Xu$^{*}$
\thanks{\dag~Contribute equally to this work, listed in random order. *~Corresponding author.}
\thanks{This work was supported by Shenzhen Ubiquitous Data Enabling Key Lab under Grant No. ZDSYS20220527171406015, by Guangdong Innovative and Entrepreneurial Research Team Program (2021ZT09L197), by Shenzhen Science and Technology Program (JCYJ20220530143013030), by Tsinghua Shenzhen International Graduate School-Shenzhen Pengrui Young Faculty Program of Shenzhen Pengrui Foundation (No. SZPR2023005).}
\thanks{C. Li, Z. Ai, T. Wu, X. Li and W. Ding are with Tsinghua-Berkeley Shenzhen Institute, Tsinghua Shenzhen International Graduate School, Tsinghua University, Shenzhen, China. E-mail: {\tt\small \{li-cc21, azh21, wu-t23, lixs21\}@mails.tsinghua.edu.cn} and {\tt\small ding.wenbo@sz.tsinghua.edu.cn}}
\thanks{W. Ding is also with RISC-V International Open Source Laboratory, Shenzhen, China, 518055.}
\thanks{H. Xu is with Institute for Interdisciplinary Information Sciences (IIIS), Tsinghua University, Beijing, China. E-mail: {\tt\small huazhe$\_$xu@mail.tsinghua.edu.cn}}
}

\begin{document}

\maketitle
\thispagestyle{empty}
\pagestyle{empty}

\begin{abstract}
    Manipulating deformable objects is a ubiquitous task in household environments, demanding adequate representation and accurate dynamics prediction due to the objects' infinite degrees of freedom. This work proposes DeformNet, which utilizes latent space modeling with a learned 3D representation model to tackle these challenges effectively. The proposed representation model combines a PointNet encoder and a conditional neural radiance field~(NeRF), facilitating a thorough acquisition of object deformations and variations in lighting conditions. To model the complex dynamics, we employ a recurrent state-space model~(RSSM) that accurately predicts the transformation of the latent representation over time. Extensive simulation experiments with diverse objectives demonstrate the generalization capabilities of DeformNet for various deformable object manipulation tasks, even in the presence of previously unseen goals. Finally, we deploy DeformNet on an actual UR5 robotic arm to demonstrate its capability in real-world scenarios.
\end{abstract}

\input{tables/1intro}
\input{tables/2relatedwork}

\input{tables/3method}

\input{tables/4experiment}

\input{tables/5conclusion}
\bibliographystyle{IEEEtran}

\balance
\bibliography{ref}
\addtolength{\textheight}{-12cm}   




\end{document}

%% file: tables/1intro.tex
\section{Introduction} 
    \begin{figure}[htbp]
		\centering
		\includegraphics[width=\linewidth]{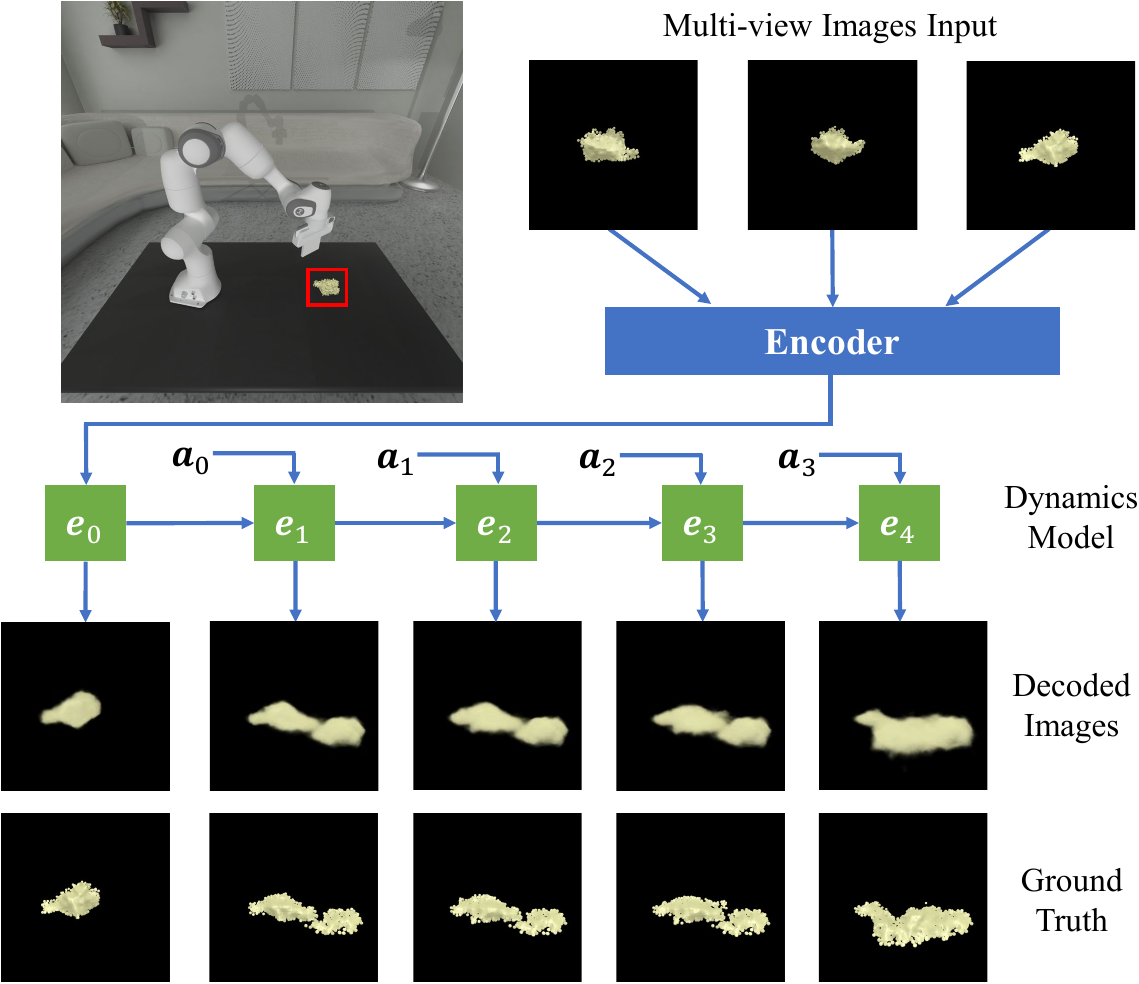}
		\caption{\textbf{Overview of DeformNet framework.} For a given time step, an encoder initially transforms images captured from diverse viewpoints into a coherent embedding. Subsequently, a dynamics model is employed to forecast the resultant embedding following a set of sampled actions. Finally, a NeRF decoder is utilized to generate images conditioning on the forecasted embeddings.}
		\label{fig:exptraj}
	\end{figure}
    Manipulating deformable objects is crucial in robotics given its ubiquity in everyday activities, spanning from molding dough to tying knots in ropes. 
    Previous research has proposed specific representations and techniques for different deformable objects, such as ropes~\cite{antonova2022bayesian,lee2021sample,yan2021learning}, cables~\cite{sanchez2020tethered,seita2021learning,she2021cable}, clothes~\cite{ha2022flingbot,lin2022learning,wu2019learning}, fluid~\cite{legaard2023constructing,li20223d}, plasticine~\cite{sanchez2018robotic,li2018learning, matl2021deformable}, clay~\cite{gu2023maniskill2}, and gauze~\cite{thananjeyan2017multilateral}.  
    We instead ask the question: Can we model and manipulate such objects using only RGB-D observations without explicit inductive bias about physical properties, such as particle positions and mass distributions?
    
    We argue that discovering a general state representation is at the heart of deformable object manipulation.
    Highly deformable objects, such as dough, pose challenges in capturing detailed deformations, as their shapes can exhibit significant variations. Hence, traditional convolutional neural networks~(CNNs), which primarily capture 2D features, are unable to model these highly deformable objects. 
    In recent years, neural radiance fields~(NeRF)~\cite{mildenhall2020nerf} have gained attention for their ability to precisely reconstruct 3D scenes even when the scenes are dynamic~\cite{driess2023learning, li20223d}, making it suitable for various control tasks. 

    With a reasonable representation in hand, another hurdle lies in the prediction of the dynamics, particularly when deformation is large. At first glance, some readers may wonder whether model-free reinforcement learning (RL) can circumvent the prediction by optimizing a manipulation policy directly. However, RL approaches usually struggle to find suitable reward functions and optimize actions in the face of high DoFs of deformable objects~\cite{lin2021softgym}.
    Recently, graph neural networks~(GNNs) have been proposed to model the long-term dynamics of soft materials~\cite{li2018learning, driess2023learning,shi2022robocraft}.
    While GNNs have shown great promise in long-term prediction, their reliance on neighboring relations makes it difficult to effectively encode information such as rotations. Additionally, GNNs also require a predefined underlying structure, and the task of adapting these networks to dynamic graphs continues to present a research challenge.
    Some prior works have attempted latent space modeling for soft body manipulation~\cite{yan2021learning, li20223d} with simple multilayer perceptron~(MLP) as the dynamics model; however, these works have limited performance for objects with high DoFs. We train a recurrent state-space model (RSSM)~\cite{hafner2019learning} as a world model to learn the complex dynamics of deformable objects, which can simultaneously capture global features as well as internal relations.
    
    In this work, we propose a simple and unified framework named DeformNet which combines a learned world model with a conditioned NeRF structure to address the aforementioned challenges. More specifically, we first use a PointNet encoder to extract latent vectors from the point cloud converted from RGB-D images. Then, the encoded latent vectors are split into latent deformation vectors and latent appearance vectors, where the former are responsible for density predictions, and the latter are for prediction of RGB colors. In this way, our NeRF structure is capable of representing complex 3D shapes, which is essential for the world model to accurately predict the dynamics. Finally, we add the gradient descent step to sample efficient cross-entropy method~\cite{pinneri2021sample} for trajectory planning, using reward prediction from the world model as the cost function. \ours achieves outstanding performance on complex manipulation tasks with various targets, including pinching plasticine to different shapes, writing on the clay, and towel manipulation.  
    
    The main contributions of this paper are tri-fold: 
    \begin{itemize}
        \item We propose DeformNet, which augments the neural radiance field with a latent deformation vector and a latent appearance vector, enabling the accurate representation of largely deformed objects. 
        \item By incorporating gradient-based planning with a learned world model, \ours exhibits substantial performance on complex manipulation tasks. 
        \item \ours also showcases stability and generalization ability, demonstrating its potential for real-world applications in deformable object manipulation across various tasks and shapes.
    \end{itemize}

%% file: tables/2relatedwork.tex
\section{Related Works}
    
\textbf{Representations of deformable objects.} In recent years, data-driven methods have drawn lots of attention to represent deformable shapes in low-dimensional latent space. Latent vectors~\cite{kurutach2018learning,yan2021learning,chen2022diffsrl,lippi2020latent}, keypoint embeddings~\cite{li2020causal,ma2021learning} and mesh embeddings~\cite{tan2020realtime} have been used to represent deformable objects such as ropes~\cite{ma2021learning,kurutach2018learning,yan2021learning}, clothes~\cite{li2020causal,tan2020realtime,lippi2020latent} and plasticine~\cite{chen2022diffsrl}. Recently, neural radiance fields have been studied in non-rigid object representation due to its high-fidelity 3D reconstruction capability. Li et al.~\cite{li20223d} leverage contrastive learning and neural radiance fields to learn latent representations of pouring scene. Driess et al.~\cite{driess2023learning} propose to use a 3D feature encoder to obtain the latent vector of a rope and reconstruct the rope with learned conditional neural radiance fields. Park et al.~\cite{park2021nerfies} use a deformation field to model the transformation of complex facial expressions. However, existing works focus on representing only one type of deformable objects. By contrast, we explore the capability of neural radiance fields in representing different types of deformable objects such as ropes, plasticine, grains, and clay, with PointNet~\cite{qi2017pointnet} as encoder.

\textbf{Deformable object manipulation.} The manipulation of deformable objects has been extensively studied, with various approaches proposed in the literature. Some works have focused on adaptive methods~\cite{navarro2016automatic,cherubini2020model,yoshimoto2011active}, while others have explored simulation-based approaches for actual manipulation~\cite{ganapathi2020learning}. However, the challenge of simulation-based method lies in accurately estimating the parameters, which creates a barrier in bridging the sim2real gap. Imitation learning is also leveraged to shape sand~\cite{cherubini2020model} or dough~\cite{figueroa2016learning}, but obtaining demonstrations for training can be expensive.  In recent developments, model-based methods~\cite{cretu2011soft,li2022contact,lindiffskill} have gained increasing attention. Matl and Bajcsy~\cite{matl2021deformable} utilize bounding box to represent dough and shape it to different lengths. Shi et al.~\cite{shi2022robocraft} propose GNN-based dynamics and deform plasticine to more complex character shapes. Unlike methods utilizing manually designed features or numerous particles to represent deformable objects, \ours employs latent vectors. To accurately predict the deformation of complex shapes in latent space, we utilize RSSM~\cite{hafner2019learning} as the underlying dynamics model. 

%% file: tables/3method.tex
\section{Method}
\begin{figure}
		\centering
		\includegraphics[width=1\linewidth]{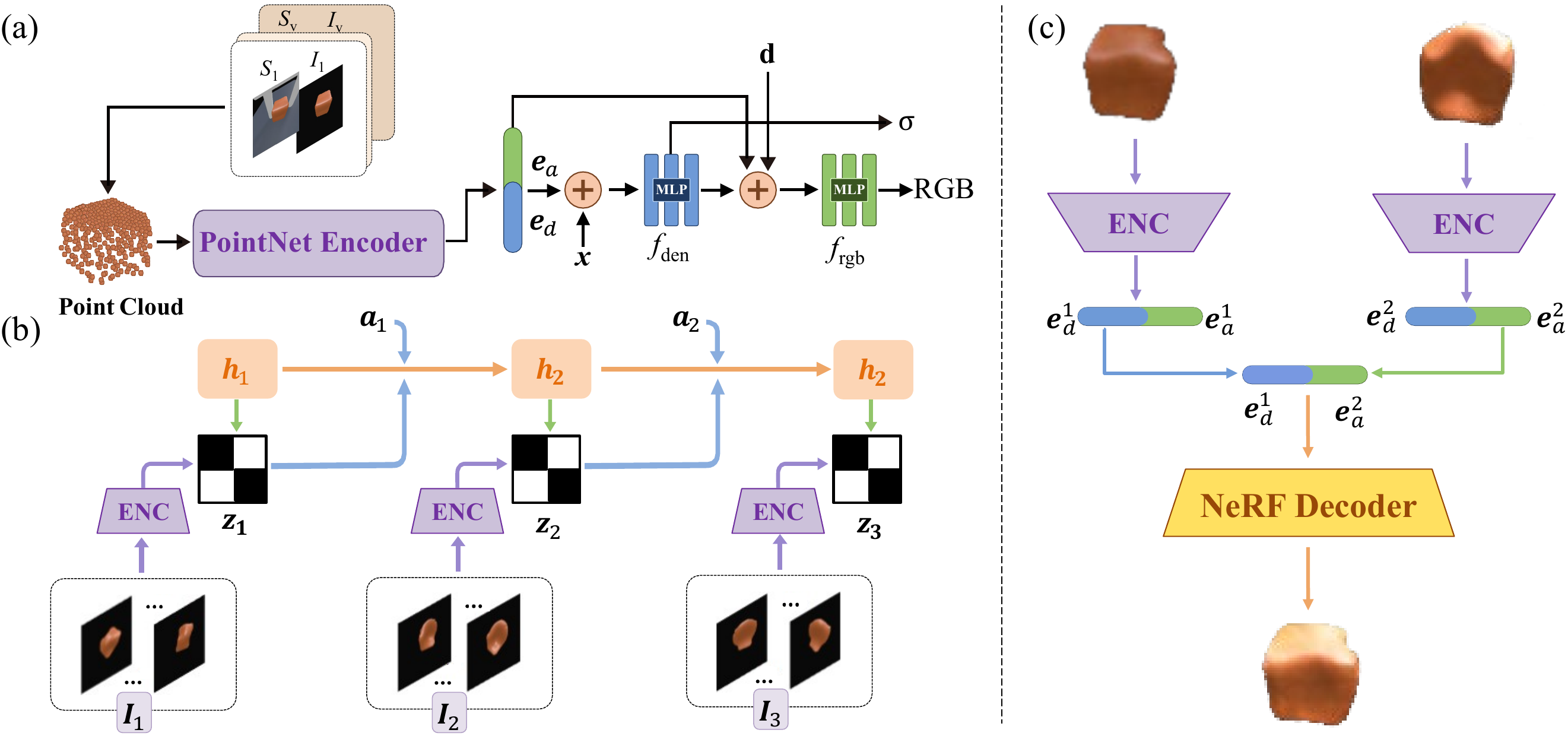}
		\caption{\textbf{Overview of the model architecture.} (a) The PointNet encoder generates latent features. These latent features are split and utilized to enhance the NeRF. (b) We then employ RSSM to capture the dynamics within the trained latent space. (c) We examine the effect of latent deformation vectors and latent appearance vectors in a fixed dataset with various lighting conditions and object shapes.}
		\label{fig:modeloverview}
\end{figure}
\subsection{Problem Formulation}

    Our research focuses on the utilization of a variety of actuators to manipulate or reposition a deformable object to a predefined position or configuration. Given the initial observation $s_0$ of the object and the target observation $s_g$, our robotic system needs to execute a sequence of actions $\mathbf{a}_{0,...,H-1}\in \mathcal{A}$ to bring the observation $s_{H}$ close to the desired goal observation $s_g$.

    To accomplish this, we employ the recurrent state space model (RSSM)~\cite{hafner2019learning} denoted as $\Phi$ to learn the transition model $\Phi:\mathcal{S}\times\mathcal{A}\rightarrow\mathcal{S}$ of the deformable object. At each time step $t$, the RSSM takes the current environmental observations $s_t$ and a sequence of actions $\mathbf{a}_{t,...,t+H-1}$, and predicts future observations $s_{t+1,...,t+H}$, with $H$ denoting the length of the prediction horizon. This predictive model allows us to formulate our manipulation tasks as a model predictive control (MPC) problem. Considering the range of our goals, we adopt different cost functions to gauge the distance between the object's current observation and the goal observation. The action sequence to be carried out is chosen based on the cost function:

\begin{equation}\label{pf}
\left(\mathbf{a}_0, \!\ldots, \!\mathbf{a}_{H\!-\!1}\!\right)\!=\!\!\underset{\mathbf{a}_{0 \ldots H-1} \in \mathcal{A}}{\!\!\!\arg \min }\!\!\! \mathcal{J}\!\left({\Phi}\!\left(s_0,\!\left(\mathbf{a}_0, \ldots, \mathbf{a}_{H\!-\!1}\right)\right)\!, \!s_g\right)\!.
\end{equation}

\subsection{3D Representation Learning for Deformable Objects}
\ours utilizes the encoder-decoder framework for 3D representation learning, as shown in Fig. \ref{fig:modeloverview}(a).
    
   \textbf{PointNet encoder.} We use the PointNet~\cite{qi2017pointnet} architecture as the encoder to capture the overall features of the object, as illustrated in Fig. \ref{fig:modeloverview}(a). RGB-D image observation $s_t$ from different viewpoints at time step $t$ are converted into a single point cloud $\mathbf{P} = (\mathbf{c}_p^{n\times3}, \mathbf{x}_p^{n\times3})$ using camera parameters, where $\mathbf{c}_p^{n\times3}$ is the RGB color and $\mathbf{x}_p^{n\times3}$ is the 3D location. Subsequently, the PointNet encoder $f_{\omega}$ processes the input point cloud to generate an embedding $\mathbf{e} = f_{\omega}(\mathbf{c}_p^{n\times3}, \mathbf{x}_p^{n\times3})$. This embedding is then divided into two parts: a latent deformation vector $\mathbf{e}_d$ and a latent appearance vector $\mathbf{e}_a$. As detailed below, $\mathbf{e}_d$ contains the shape information and $\mathbf{e}_a$ denotes the lighting condition.

    \textbf{Neural radiance field.} With 3D coordinate $\mathbf{x}\subseteq{\mathbb{R}^3}$ and viewing direction unit vector $\mathbf{d}\subseteq{\mathbb{R}^3}$ as input, NeRF learns a function that maps spatial location orientation to its RGB color value $\mathbf{c}(\mathbf{x},\mathbf{d})$ and volume density $\sigma(\mathbf{x})$~\cite{mildenhall2020nerf}. The color of an image pixel from a particular viewpoint can be rendered based on volumetric rendering function:
    \begin{small}
    \begin{equation}
        \hat{\boldsymbol{C}}(\mathbf{r})\!=\!\!\int_{\alpha_n}^{\alpha_f} \!\!T(\alpha) \sigma(\alpha) \boldsymbol{c}(\alpha) d\alpha\!,\!\ \
        T(\alpha)\!=\!\exp \left(\!-\!\int_{\alpha _n}^\alpha \!\!\sigma(s) d s\right),
    \end{equation}
    \end{small}
    where $\alpha _n$ and  $\alpha _f$ are near bound and far bound, $\mathbf{r}=\mathbf{o}+\alpha \mathbf{d}$ is the camera ray with origin $\mathbf{o}\in\mathbb{R}^3$ and the viewing direction $\mathbf{d}$, and $T(\alpha)$ denotes the accumulated transmittance along the ray from $\alpha _n$ to $\alpha_f$. During the training phase, NeRF is optimized by $\ell_2$ loss $\mathcal{L}_{\text{rec}}$ between the reconstructed color $\hat{\boldsymbol{C}}$ and ground truth color $\boldsymbol{C}$.

    \textbf{Conditional NeRF decoder.} In the standard NeRF formulation, one model is trained to render one static scene. It has been proposed in the literature to use latent vectors as conditioned inputs to NeRF~\cite{22-driess-NeRF-RL, park2021nerfies}. However, previous works mostly used NeRF trained on conditioned inputs to render a scene where the variation is limited~\cite{park2021nerfies} or the objects are composed of rigid bodies~\cite{22-driess-NeRF-RL}.

    Our task setting presents a key challenge as the deformable object may deform severely after applying a sequence of actions. Accurately manipulating highly deformable materials requires adequate 3D representation of the objects. When the object expands, it usually can be easily represented as this type of deformation causes changes in pixels. 
    However, it is difficult to capture the changes in the dented surface during deformation, as the dented area may only manifest as a slight darkening of the pixels. Instead of using a single latent vector as input for NeRF, we use latent deformation vector $\mathbf{e}_d$ to represent the shape change and latent appearance vector $\mathbf{e}_a$ to capture the lighting condition.  As in Fig. \ref{fig:modeloverview}(a), we utilize $\sigma =f_{\text{den}}(\mathbf{x},\mathbf{e}_d)$ to predict density and $\mathbf{c}=f_{\text{rgb}}(\mathbf{F},\mathbf{e}_a, \mathbf{d})$ to predict color. Here, $\mathbf{F}$ represents the 256-dimensional feature generated by $f_{\text{den}}$. 
    If we have two observations $s_1$ and $s_2$ with different shapes and lighting conditions, we can combine the deformation latent $\mathbf{e}_d^1$ encoded from $s_1$ with the appearance latent $\mathbf{e}_a^2$ from $s_2$, which is illustrated in Fig. \ref{fig:modeloverview}(c). Subsequently, as we feed this resultant latent representation into the NeRF decoder, we observe a transfer of the lighting conditions from $s_2$ to $s_1$.
    


\subsection{Learning Dynamics of Deformable Object via RSSM}

RSSM~\cite {hafner2021mastering} is a recurrent model that incorporates deterministic states $\mathbf{h}_t$ and categorical stochastic states $\mathbf{z}_t$ for long-term predictions. As demonstrated in Fig. \ref{fig:modeloverview}(b), the components of our predictive model are as follows:
    \begin{equation}
        \begin{aligned}
        &\text{Recurrent model:} &\mathbf{h}_t &=f_\phi\left(\mathbf{h}_{t-1}, \mathbf{z}_{t-1}, \mathbf{a}_{t-1}\right) \\
        &\text{Representation model:} &\mathbf{z}_t &\sim q_\phi\left(\mathbf{z}_t \mid \mathbf{h}_t, \mathbf{e}_t\right) \\
        &\text{Reward predictor:} &\hat{r}_t  &\sim p_\phi\left(\hat{r}_t \mid \mathbf{h}_t, \mathbf{z}_t, \mathbf{e}_g\right) \\
        &\text{Transition predictor:} &\hat{\mathbf{z}}_t  &\sim p_\phi\left(\hat{\mathbf{z}}_t \mid \mathbf{h}_t\right) \\
        &\text{Embedding predictor:} &\hat{\mathbf{e}}_t  &\sim p_\phi\left(\hat{\mathbf{e}}_t \mid \mathbf{h}_t, \mathbf{z}_t\right),
        \end{aligned}
    \end{equation}
where $\mathbf{e}_g$ and $\mathbf{e}_t$ are the latent vectors encoded from goal images and current observation respectively, and $\phi$ describes the mutual parameters. At each step $t$, the RSSM computes the posterior stochastic state $\mathbf{z}_t$, incorporating information about the current embedding $\mathbf{e}_t$ obtained from the PointNet encoder. It also predicts a reconstructed embedding $\hat{\mathbf{e}}_t$, and a prior stochastic state $\hat{\mathbf{z}}_t$ that anticipates the posterior without access to the embedding. The reward predictor predicts the opposite value of the cost functions.
The training process involves minimizing the embedding reconstruction loss, goal-conditioned reward log loss, and the KL loss of posterior and prior distributions. The loss function is defined as:
\begin{align}
\mathcal{L}_{\text{dym}}=\sum_{t=0}^{H-1}\left\|\mathbf{e}_t-\hat{\mathbf{e}}_t\right\|_2^2-\ln p_{\phi}\left(r_t \mid \mathbf{h}_t, \mathbf{z}_t, \mathbf{e}_{g}\right) \nonumber\\ 
+\text{KL}\left(q_{\phi}\left(\mathbf{z}_t \mid \mathbf{h}_t, \mathbf{e}_t\right) \| p_{\phi}\left(\hat{\mathbf{z}}_t \mid \mathbf{h}_t\right)\right).
\end{align}
During training, we first train the PointNet encoder and NeRF decoder by optimizing $\mathcal{L}_{\text{rec}}$, then we freeze the parameters of the encoder-decoder framework and train RSSM by minimizing $\mathcal{L}_{\text{dym}}$. To avoid leading to a poorly trained prior, we also employ KL-balancing~\cite{hafner2021mastering}. This approach allows us to model the dynamics of the system and predict future states, providing a robust and efficient solution for target tasks.

\subsection{Cost Functions} 
To quantify the spatial similarity between the distributions of MPM particles, we employ the following cost functions for our simulation environments including the Chamfer distance (CD), the earth mover's distance (EMD)~\cite{shi2022robocraft}, and the soft intersection over union (SIoU)~\cite{huang2021plasticinelab}. We also adopt the discrete-to-continuous distance (D2CD). The D2CD is adapted from the average Chamfer distance~\cite{suh2021surprising}, which gauges the distance between a discrete set of MPM particles and a continuous target set, denoted as $\mathcal{O}$ and $\mathcal{S}_d$ in $\mathbb{R}^3$. It is defined as follows:
    \begin{equation}
        \mathcal{J}_{\mathrm{D2CD}}\left(\mathcal{O}, \mathcal{S}_d\right)=\frac{1}{|\mathcal{O}|}\sum_{\mathbf{x} \in \mathcal{O}} \min _{\mathbf{y} \in \mathcal{S}_d}\|\mathbf{x}-\mathbf{y}\|_2.
    \end{equation}

\subsection{Closed-Loop Control}\label{icem}
After obtaining the encoder and the dynamics model. We aim to optimize an action sequence $\mathbf{a}_{0,...,H-1}$ in a designed action space. This sequence should make the final observation $s_{H}$ as close as possible to the desired observation $s_{g}$, as shown in Equation~\eqref{pf}. To achieve this, we adopt the sample efficient cross-entropy method (iCEM)~\cite{pinneri2021sample}, which utilizes the gradient generated by our predictive model to improve the selected actions. Intermediate observations are also used to correct model predictions as in standard MPC setting~\cite{shi2022robocraft}. This planning process continues until either the predicted reward exceeds a predefined threshold or the action sequence reaches its maximum length.
   
During the closed-loop control, the cameras capture images of the workspace from various angles. These images are used as observations and to create a point cloud for the training of encoder and decoder. Our method takes as input the current observation and the final target, and calculates the best actions to achieve the goal. The robot interacts with the deformable object, adjusts its position to ensure a clear view for the cameras, and then plans its next move based on the visual feedback.

%% file: tables/4experiment.tex
\section{Experiments}
\label{sec:result}
\subsection{Tasks and Environments}
 \begin{figure}
		\centering
		\includegraphics[width=\linewidth]{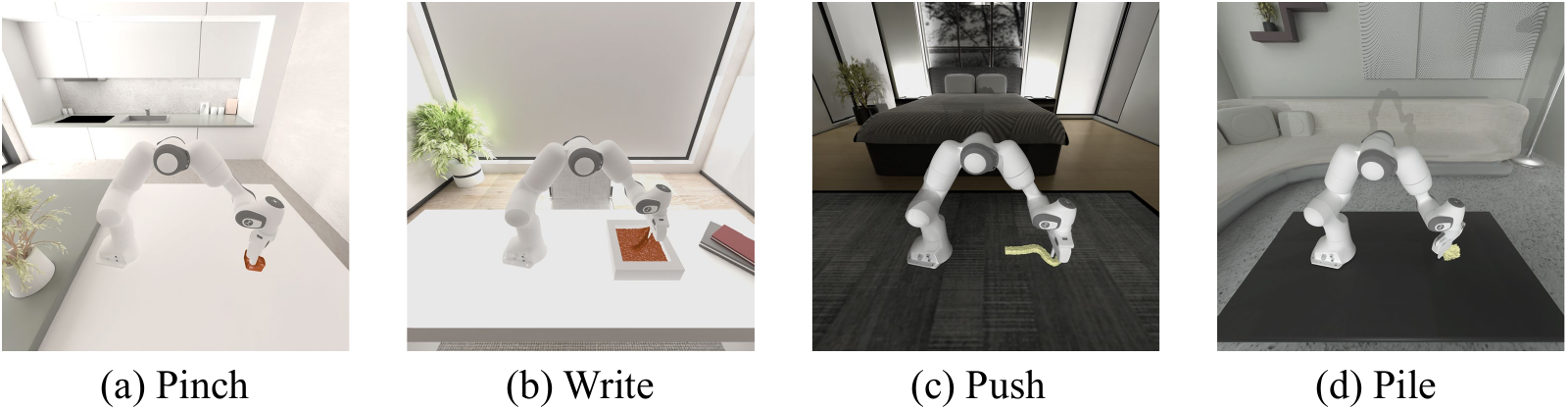}
		\caption{\textbf{Simulator visualizations.} (a) The robot demonstrates its ability to manipulate plasticine through pinching actions, corresponding to the environment 'Pinch'. In the 'Poke' task, the gripper is replaced with a stick. (b) The robot showcases its writing skills by creating characters on a clay surface in environment 'Write'. (c) In the 'Push' task, the robot manipulates the towel by pushing it with a stick to achieve specific target shapes. (d) Within the 'Transport' and 'Pile' environments, the robot adeptly either relocates grains to desired locations or pushes them into predefined forms. }
		\label{fig:env}
	\end{figure}
 
  We implement our simulation environment based on Maniskill2~\cite{gu2023maniskill2}, an MPM-based soft-body manipulation environment that facilitates real-time simulation, as visualized in Fig. \ref{fig:env}. The performance of \ours is evaluated on six challenging deformable manipulation tasks namely 'Push', 'Pinch', 'Write', 'Pile', 'Transport', and 'Poke'. The 'Push' task involves pushing a soft towel on a 2D plane using a robot equipped with a stick. In the 'Pinch' task, the environment features a robot equipped with a gripper that manipulates plasticine into several predefined target shapes. For the 'Write' task, the robot employs a stick to write some characters on the clay surface. In the 'Pile' task, the robot's challenge is to pile grains on a table into specific target shapes using a board. Similar to the 'Pile' task, the 'Transport' task involves the same scenario but focuses on relocating grains on the table. Lastly, in the 'Poke' task, a robot with a stick prods plasticine into some target shapes. For simplicity, we assume a deterministic initial observation for all tasks. In the simulation experiments, all tasks are executed using a 7 DoFs Franka Emika robot with 8 cameras around the deformable object. 

\begin{figure}
		\centering
		\includegraphics[width=\linewidth]{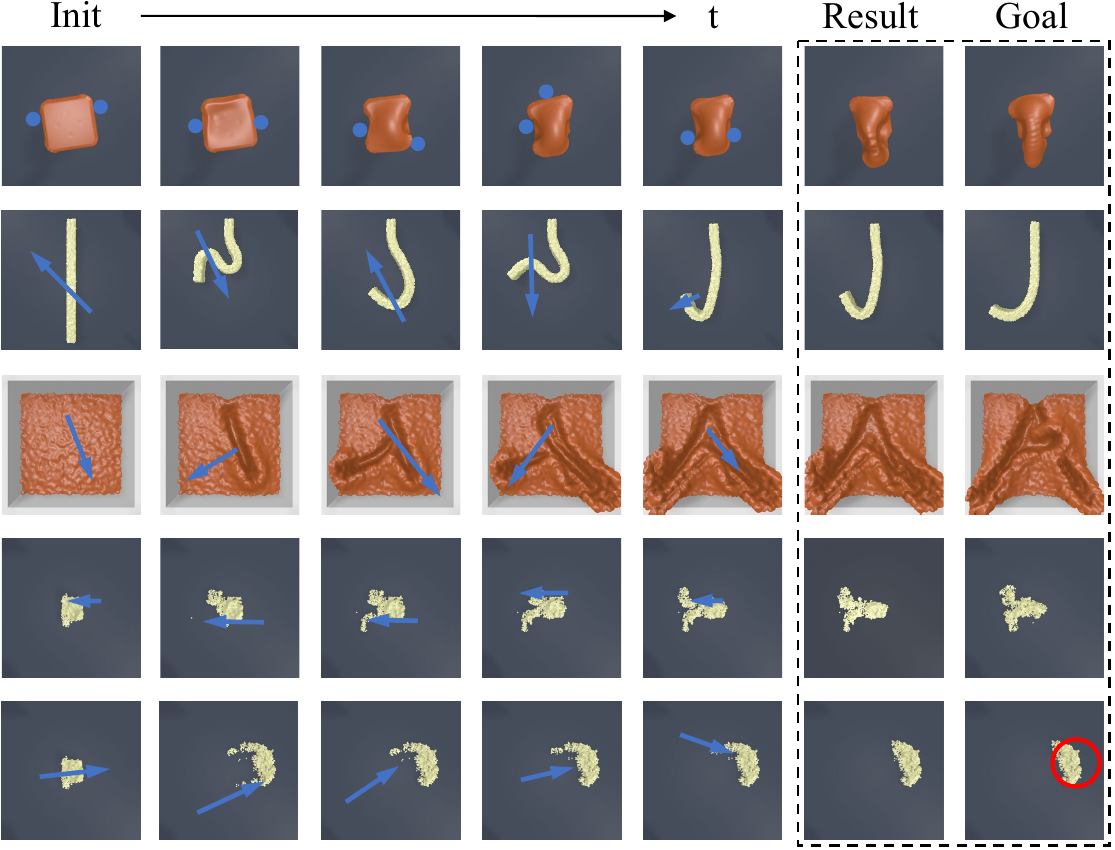}
		\caption{\textbf{Control results and example trajectories by DeformNet.} The target configuration to be achieved is depicted in the image on the right. The planning process is illustrated in the left five columns, while the control results obtained from the algorithm are displayed in the sixth column. The five rows correspond to the 'Pinch', 'Push', 'Write', 'Pile', and, 'Transport' tasks, respectively. In the 'Pinch' task, the gripper's action is symbolized by a blue dot. In the 'Push' and 'Write' tasks, the actions executed by the stick are illustrated by blue arrows. In the 'Pile' and 'Transport' tasks, the arrows represent the piling direction and distance.}
		\label{fig:traj}
	\end{figure}

\begin{table*}[!htb]

\centering
 \resizebox{\textwidth}{!}{
\begin{tabular}{l|ccc|ccc|cl}
\toprule
                                     & \multicolumn{3}{c|}{\textbf{Push}}                                                                                                   & \multicolumn{3}{c|}{\textbf{Pinch}}
                                     & \multicolumn{1}{c}{\textbf{Write}}\\
\multicolumn{1}{l|}{}                & \textbf{CD}$\downarrow$       & \textbf{EMD}$\downarrow$         & \textbf{SIoU}$\uparrow$        & \textbf{CD}$\downarrow$        & \textbf{EMD}$\downarrow$           & \textbf{SIoU}$\uparrow$            & \textbf{SIoU}$\uparrow$          \\ \midrule
\textbf{Random Policy}               & 13.28 $\pm$ 3.95                   & 4.91 $\pm$ 2.50                   & 0.13 $\pm$ 0.12                   & 3.14 $\pm$ 0.55                   & 1.27 $\pm$ 0.22              & 0.06 $\pm$ 0.08                   & 0.19 $\pm$ 0.07                  \\
\textbf{PPO}                 & 12.81 $\pm$ 3.79          & 3.76 $\pm$ 3.27          & 0.15 $\pm$ 0.09 &2.53 $\pm$ 0.15          & 1.04 $\pm$ 0.03        & 0.15 $\pm$ 0.08          & 0.19 $\pm$ 0.04           \\
\textbf{NeRF-RL}                      & 9.50 $\pm$ 2.02          & 2.91 $\pm$ 1.29          & 0.12 $\pm$ 0.05         & 2.50 $\pm$ 0.13          & 1.01 $\pm$ 0.03        & 0.26 $\pm$ 0.04          & 0.17 $\pm$ 0.06          \\
\textbf{NeRF-dy} & 11.47 $\pm$ 2.55          & 3.09 $\pm$ 0.77          & 0.19 $\pm$ 0.13          & 2.25 $\pm$ 0.43          & 0.95 $\pm$ 0.20        & 0.21 $\pm$ 0.11          & 0.18 $\pm$ 0.07          \\
\textbf{Dreamer-V2}        & 9.78 $\pm$ 0.01          & 2.75 $\pm$ 0.02 & 0.18 $\pm$ 0.09          & 2.53 $\pm$ 0.32          & 1.04 $\pm$ 0.07        & 0.22 $\pm$ 0.07 & 0.26 $\pm$ 0.06          \\
\textbf{DeformNet (Ours)}                  & \textbf{6.86} $\pm$ \textbf{3.11} & \textbf{2.66} $\pm$ \textbf{1.45}          & \textbf{0.27} $\pm$ \textbf{0.12}          & \textbf{1.87} $\pm$ \textbf{0.27} & \textbf{0.80} $\pm$ \textbf{0.13} & \textbf{0.32} $\pm$ \textbf{0.09}          & \textbf{0.53} $\pm$ \textbf{0.07}\\
\bottomrule 
\end{tabular}

}

\end{table*}

\begin{table*}

\centering
 \resizebox{\textwidth}{!}{
\begin{tabular}{l|ccc|ccc|cl}
\toprule
                                     & \multicolumn{3}{c|}{\textbf{Poke}}                                                                                                   & \multicolumn{3}{c|}{\textbf{Pile}}
                                     & \multicolumn{1}{c}{\textbf{Transport}}\\
\multicolumn{1}{l|}{}                & \textbf{CD}$\downarrow$       & \textbf{EMD}$\downarrow$         & \textbf{SIoU}$\uparrow$        & \textbf{CD}$\downarrow$        & \textbf{EMD}$\downarrow$           & \textbf{SIoU}$\uparrow$            & \textbf{D2CD}$\downarrow$          \\ \midrule
\textbf{Random Policy}               & 2.30 $\pm$ 0.15                   & 0.94 $\pm$ 0.06                   & 0.11 $\pm$ 0.05                   & 5.68 $\pm$ 2.20                   & 3.75 $\pm$ 1.80              & 0.05 $\pm$ 0.09                   & 0.11 $\pm$ 0.03                  \\
\textbf{PPO}                 & 2.28 $\pm$ 0.06          & 0.91 $\pm$ 0.01          & 0.08 $\pm$ 0.04 &4.68 $\pm$ 1.94          & 2.69 $\pm$ 1.35       & 0.05 $\pm$ 0.04          & 0.10 $\pm$ 0.01           \\
\textbf{NeRF-RL}                      & 2.26 $\pm$ 0.10          & 0.90 $\pm$ 0.04          & 0.11 $\pm$ 0.07         & 3.83 $\pm$ 1.58          & 2.46 $\pm$ 1.24        & 0.06 $\pm$ 0.05          & 0.10 $\pm$ 0.01          \\
\textbf{NeRF-dy} & 2.05 $\pm$ 0.06          & 0.81 $\pm$ 0.11          & 0.17 $\pm$ 0.08          & 3.78 $\pm$ 1.46          & 2.65 $\pm$ 1.30        & 0.13 $\pm$ 0.15          & -          \\
\textbf{Dreamer-V2}        & 1.56 $\pm$ 0.03          & \textbf{0.62} $\pm$ \textbf{0.07} & 0.27 $\pm$ 0.04          & 5.14 $\pm$ 0.61          & 4.61 $\pm$ 0.50        & 0.08 $\pm$ 0.02 & 0.10 $\pm$ 0.03         \\
\textbf{DeformNet (Ours)}                  & \textbf{1.46} $\pm$ \textbf{0.23} & 0.63 $\pm$ 0.11          & \textbf{0.45} $\pm$ \textbf{0.11}          & \textbf{2.88} $\pm$ \textbf{0.47} & \textbf{2.37} $\pm$ \textbf{0.60} & \textbf{0.20} $\pm$ \textbf{0.09}          & \textbf{0.07} $\pm$ \textbf{0.01}\\
\bottomrule 
\end{tabular}

}
\\
\caption{\upshape\textbf{Quantitative comparisons between the DeformNet and the baselines on each task.} In the 'Push', 'Pinch', 'Poke', and 'Pile' tasks, we evaluate \ours and the baselines using CD, EMD, and SIoU. Lower values of CD, D2CD, and EMD indicate better performance, while higher values of SIoU suggest better performance. Both CD and EMD are scaled by 100 in the table. In the 'Write' and 'Transport' tasks, we specifically compare the SIoU and D2CD respectively. NeRF-dy is omitted in the final column due to its reliance on the objective's latent representation, while the transport environment lacks a goal representation. \ours consistently outperforms all baselines in most testing scenarios, demonstrating its performance in generalization. }
\end{table*}\label{table:main_results}
\subsection{Action Space}
We specify the action space for each task within a parameterized framework. In tasks such as 'Push', 'Write', and 'Transport', the action space is characterized by $\{x_0,y_0,x_1,y_1\}$, where $\{x_0,y_0\}$ denotes the initial position of the end effector, and $\{x_1,y_1\}$ denotes the final destination. The motion of the stick or board from $\{x_0,y_0\}$ to $\{x_1,y_1\}$ is at a fixed height.

For the 'Pile' task, the action space is defined as $\{x_0,y_0,\delta_d\}$, where $\{x_0,y_0\}$ denotes the initial position of the board, and $\delta_d$ represents the board's movement. In the 'Poke' task, we simplify its action space to $\{x_0,y_0\}$, indicating that the robot pokes the plasticine at a consistent height.

Lastly, for the 'Pinch' task, the action is parameterized as $\{x, y, z, r_z, d_g\}$. Here, $\{x, y, z\}$ denotes the center of the gripper, $r_z$ denotes the rotation of the robot gripper along the z-axis, and $d_g$ indicates the minimal distance between the gripper fingers. We apply the inverse kinematics library from robot toolbox~\cite{rtb} and default PD control from the Maniskill2 environment.
\subsection{Baseline Methods}
    In order to provide a comprehensive comparison, we include the following baseline methods. \textbf{Random}: This baseline randomly samples actions uniformly within the action space. \textbf{PPO}: We employ the Proximal Policy Optimization (PPO)~\cite{schulman2017proximal} as the model-free RL algorithm. \textbf{NeRF-RL}: Inspired by the original NeRF-RL paper~\cite{22-driess-NeRF-RL}, we utilize a 2D convolutional neural network (CNN) to encode images into a latent representation supervised by a NeRF decoder. PPO is then applied to the learned latent space. \textbf{NeRF-dy}: This method is an adaptation of NeRF-dy~\cite{li20223d}, which omits the use of time contrastive loss during training. Instead of the model predictive path integral (MPPI) planning method, we employ iCEM as the planning approach for consistency. \textbf{Dreamer-V2}: Following the structure of Dreamer-V2~\cite{hafner2021mastering}, we adopt a 2D CNN encoder to encode images captured from different perspectives. These baseline methods serve as reference points for evaluating the performance of our proposed approach. For each baseline, we evaluate the result by sampling 100 trajectories in each task. 
\subsection{Training Details}
To train our models, we follow a specific procedure for each task. Initially, we generate 100 random trajectories and train our representation model and dynamics model separately. The dynamics model is updated with the latest weights of the trained representation model every 50 epochs.

For the PointNet encoder, we utilize RGB-D images of sizes (8, 256, 256, 4) to generate the point cloud, while RGB images of sizes (8, 128, 128, 3) are used for conditioned NeRF training. After every 1000 epochs, we employ the gradient-based iCEM to generate additional training samples. The planning phase of iCEM consists of 50 iterations. The training process is performed on 4 NVIDIA A6000 GPUs. On average, each iteration of representation training takes approximately 1 second, while dynamics training requires around 5 seconds.

We employ the stable-baseline-3 framework~\cite{JMLR:v22:20-1364} to train the PPO and NeRF-RL. Since we do not incorporate goal-conditioned input, we randomly select a single target for training and evaluation of the RL model. Additionally, we implement Dreamer-V2 using the default Actor-Critic algorithm. The Actor-Critic module, reward predictor, and terminal predictor within the Dreamer-V2 framework are conditioned on the goal-related information. In the 'Push', 'Pinch', 'Poke', and 'Pile' tasks, we use CD for training, and EMD and SIoU for evaluation. In the 'Write' task, the particle distances vary due to different writing orders for the same character. Thus, we just employ SIoU to train and evaluate the particle set's similarity to the goal. In the 'Transport' task, we only utilize D2CD rather than compare the similarity between two distributions of MPM particles. 

\subsection {Comparison with Baselines in Simulation}
Our proposed approach demonstrates effective performance in handling complex deformable manipulation tasks, as presented in Table I. We first observe that NeRF-RL surpasses the vanilla model-free RL method (PPO), indicating the benefits of utilizing a latent representation supervised by NeRF. This improvement notably enhances the learning capabilities of the robot. Moreover, Dreamer-V2 also achieves satisfactory performance in the 'Push', 'Write', and 'Poke' tasks, highlighting the capabilities of the RSSM model. However, in tasks such as 'Pinch' and 'Pile', which involve a higher degree of deformation, all RL-based algorithms, including PPO, NeRF-RL, and Dreamer-V2, exhibit relatively lower performance compared to NeRF-dy. This can be attributed to the challenges posed by the highly deformable space, where RL-based algorithms struggle to achieve high rewards and may become trapped in local minima. In contrast, NeRF-dy, which utilizes iCEM for trajectory sampling with the learned model, achieves superior performance. \ours outperforms or closely matches all baseline methods across all metrics in the six challenging tasks.
\begin{figure}
		\centering
		\includegraphics[width=0.95\linewidth]{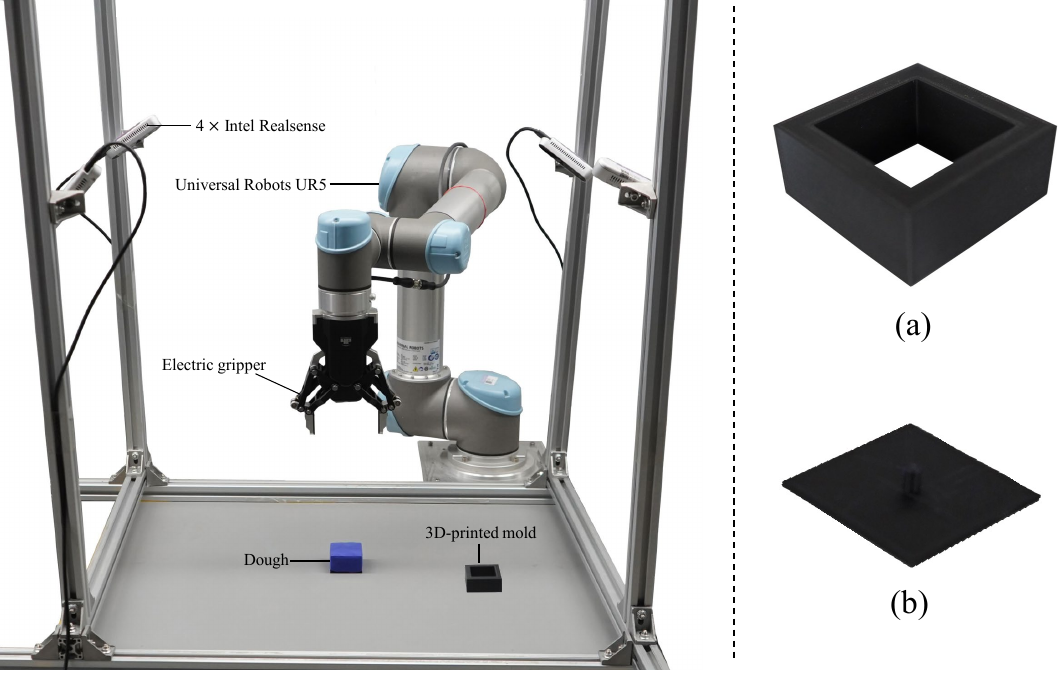}
		\caption{\textbf{Real world experiment setup.} Left: an overview of the real robot experiment setup. Right: 3D-printed tools including (a) the mold for resetting, and (b) the platform to fix the dough.}
		\label{fig:real_world}
  \vspace{-0.3cm}
	\end{figure}

\subsection{Real-World Experiment}
To evaluate the effectiveness of \ours in real-world robot systems, we conducted experiments utilizing a Universal Robots UR5 robot with 6 DoFs, as depicted in Fig. ~\ref{fig:real_world}. The robot is connected to a computer with a NVIDIA 3080 GPU via socket communication. We control the electric gripper through the I/O interface. The experiment setup involves 4 Intel RealSense D415 RGB-D cameras at different positions around the dough. These cameras are calibrated and capture RGB-D images at a resolution of 1280$\times$720 and a frame rate of 30 Hz. The initial shape of the plasticine is approximately a cuboid measuring 5$\times$5$\times$3 $\text{cm}^3$. Our goal is to manipulate the dough into the shape of the letter 'H' or 'X'. To ensure consistent starting conditions, we insert the dough into a 3D-printed mold, shaped as a cuboid cavity at the beginning of each episode.

In our real world experiment, we simplify the bounded action space into the parameterized set $\{x, y, z, r_z, d_g\}$. For data collection, we randomly collect about 300 pinching episodes for each step in the bounded action space. Then, we convert the RGB-D images captured during these episodes into point clouds with the camera parameters. To enhance the quality of the data, we apply the RANSAC algorithm~\cite{fischler1981random} to remove the outliers in the point cloud data. For training the NeRF decoder, we set near bound $\alpha_n=0.1$ and far bound $\alpha_f=0.4$. Fig.~\ref{fig:real_result} demonstrates that the resulting pinching behavior closely matches the desired shape. To provide a quantitative assessment, we executed a total of 50 trials after training the model. Among these, 36 trials yield a Chamfer distance measurement of less than 1.3 cm, corresponding to a success rate of 72$\%$. Further quantitative verification is presented in Table II, affirming that \ours consistently achieved the lowest Chamfer distance and the Earth mover's distance. This real-world experiment validates the effectiveness and practical applicability of our proposed approach in deformable robotic manipulation scenarios.

%% file: tables/5conclusion.tex
\section{Conclusion}
This study presents DeformNet, a novel approach to learn representations of deformable objects from visual observations by employing an autoencoding framework enriched with a neural radiance field rendering module. In addition, we leveraged RSSM to capture the dynamics and facilitate accurate prediction. The effectiveness of DeformNet was demonstrated through its generalization capabilities across diverse deformable manipulation tasks. Experimental evaluations confirmed the superiority of our system compared to multiple baseline methods. Furthermore, our system demonstrated compatibility with an iCEM planner, emphasizing its practical application in real-world scenarios. The successful combination of deformable representations, dynamics learning and control provides a promising foundation for the manipulation of deformable objects.
\begin{figure}
		\centering
		\includegraphics[width=0.95\linewidth]{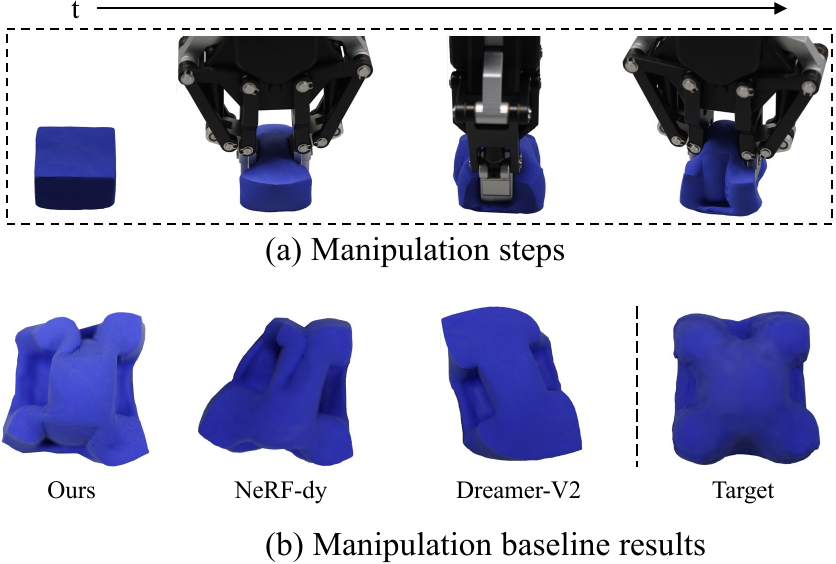}
		\caption{\textbf{Manipulation results.} (a) The robot successfully molds the dough into the shape of the letter 'X'. (b) Left: the manipulation results of three baseline methods. Right: the target shape for reference.}
		\label{fig:real_result}
	\end{figure}

\begin{table}
\centering

\begin{tabular}{lccccccc}

\toprule
\multicolumn{1}{l}{}                & \textbf{CD} (cm) $\downarrow$       & \textbf{EMD} (cm) $\downarrow$         & \textbf{Human evaluation}$\uparrow$       \\ \midrule
\textbf{NeRF-dy}               & 1.28 $\pm$ 0.11                   & 0.76 $\pm$ 0.04                   & 0.71 $\pm$ 0.16                  \\
\textbf{Dreamer-V2}               & 1.34 $\pm$ 0.04                   & 0.79 $\pm$ 0.03                   & 0.48 $\pm$ 0.07                  \\
\textbf{DeformNet}               & \textbf{1.14} $\pm$ \textbf{0.09}                   & \textbf{0.69} $\pm$ \textbf{0.04}                   & \textbf{0.83} $\pm$ \textbf{0.14}                  \\

\bottomrule 

\end{tabular}
\caption{\textbf{Quantitative comparisons between the DeformNet and the baselines in the real world scenario.} In our evaluation, we compare the performance of DeformNet with baseline methods using metrics such as CD and EMD. For the human evaluation, we invite 50 humans to evaluate our results. The results highlight the superior performance of DeformNet.}
\vspace{-0.4cm}
\end{table}